\documentclass[letterpaper, 10 pt, conference]{ieeeconf}  % Comment this line out if you need a4paper

\IEEEoverridecommandlockouts
\usepackage[table]{xcolor}
\usepackage{cite}

\usepackage{amsmath,amssymb,amsfonts}
\usepackage{graphicx}
\usepackage{textcomp}
\usepackage{authblk}
\let\labelindent\relax
\usepackage{enumitem}
\usepackage{breqn}
\usepackage{algpseudocode}
\usepackage[ruled,lined]{algorithm2e}

\usepackage{subfigure}

\usepackage{array}
\usepackage{makecell}
\usepackage{tabularray}
\usepackage{bm}

\usepackage{svg}
\usepackage{booktabs}
\usepackage{multirow}
\usepackage[colorlinks,linkcolor=red,anchorcolor=blue,citecolor=green]{hyperref}
\usepackage{threeparttable}
\usepackage{algorithm}

\title{\LARGE \bf
PP-TIL: Personalized Planning for Autonomous Driving with Instance-based Transfer Imitation Learning}
\author{
Fangze Lin$^{1}$, Ying He$^{1,*}$ and Fei Yu$^{2}$
\thanks{This work is supported in part by Shenzhen Science and Technology Program under Grant ZDSYS20220527171400002, the National Natural Science Foundation of China (NSFC) under Grants 62271324, 62231020 and 62371309, and the Guangdong Basic and Applied Basic Research Foundation under Grant 2023A1515011979.}
\thanks{$^1$College of Computer Science and Software Engineering, Shenzhen University, P.R. China. (\texttt{linfangze2023@email.szu.edu.cn}, \texttt{heying@szu.edu.cn})} 
\thanks{$^2$College of Computer Science and Software Engineering, Shenzhen University, P.R. China, and also with Carleton University, Canada. (\texttt{yufei@szu.edu.cn})}
\thanks{$^*$Corresponding Author: Ying He.} 
}

\begin{document}
\maketitle
\thispagestyle{empty}
\pagestyle{empty}
\begin{abstract}
% 1 challenge
Personalized motion planning holds significant importance within urban automated driving, catering to the unique requirements of individual users. Nevertheless, prior endeavors have frequently encountered difficulties in simultaneously addressing two crucial aspects: personalized planning within intricate urban settings and enhancing planning performance through data utilization. The challenge arises from the expensive and limited nature of user data, coupled with the scene state space tending towards infinity. These factors contribute to overfitting and poor generalization problems during model training. 
% 2 approach
Henceforth, we propose an instance-based transfer imitation learning approach. This method facilitates knowledge transfer from extensive expert domain data to the user domain, presenting a resolution to these issues. We initially train a pre-trained model using large-scale expert data. Subsequently, during the fine-tuning phase, we feed the batch data, which comprises expert and user data. Employing the inverse reinforcement learning technique, we extract the style feature distribution from user demonstrations, constructing the regularization term for the approximation of user style. 
% 3 experiment
In our experiments, we conducted extensive evaluations of the proposed method. Compared to the baseline methods, our approach mitigates the overfitting issue caused by sparse user data. Furthermore, we discovered that integrating the driving model with a differentiable nonlinear optimizer as a safety protection layer for end-to-end personalized fine-tuning results in superior planning performance. 
% code
The code will be available at https://github.com/LinFunster/PP-TIL.

\end{abstract}
\section{Introduction}
% 研究课题引入
Autonomous driving has been a focal point of research and development over the past decade. The motion planning module stands out as a critical component within the urban autonomous driving system \cite{zhu2021survey,teng2023motion}. This module typically considers factors such as safety, travel efficiency, and comfort. In practical terms, individual perceptions of comfort can vary significantly among users. For instance, some may prefer high-acceleration sports driving, while others may lean towards a more relaxed style. Offering personalized motion planning holds immense significance in enhancing user acceptance of autonomous driving. However, manually adjusting the parameters of the planning model to achieve a specified style can be challenging due to the potential for antagonistic effects of a large number of parameters. Fortunately, the data-driven approaches can effectively address this issue \cite{kuderer2015learning,schrum2024maveric}.

\begin{figure}[t]
    \setlength{\abovecaptionskip}{0cm}

    \centering
    \includegraphics[width=1.0\columnwidth]{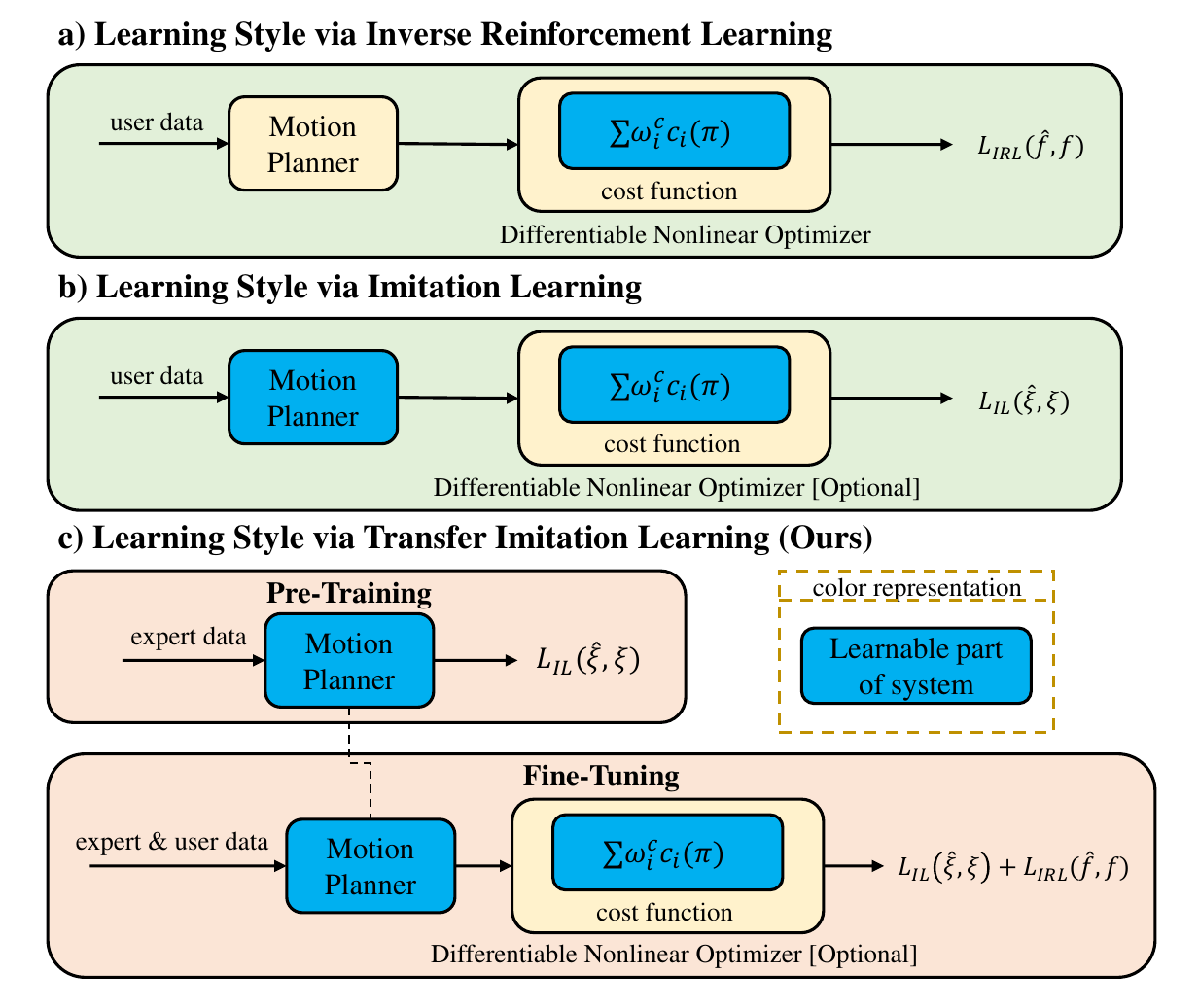}
    \vspace{-0.6cm}
    \caption{Approaches to learning driving style can generally be categorized into two main groups:
    \textbf{(a)} The first class involves using inverse reinforcement learning to learn cost functions from demonstrations. 
    \textbf{(b)} The second class involves using imitation learning to learn neural networks from demonstrations. 
    \textbf{(c)} Nevertheless, these methods often encounter overfitting and poor generalization challenges when learning from sparse user demonstrations. To address these challenges, we propose an approach termed \textbf{P}ersonalized \textbf{P}lanning via \textbf{T}ransfer \textbf{I}mitation \textbf{L}earning (PP-TIL). 
    }
    \label{intro_figure}
    \vspace{-0.5cm}
\end{figure}

For data-driven personalization planning, Fig. \ref{intro_figure} presents a comparative analysis between our proposed approach and prior methodologies. Existing approaches can be broadly categorized into two main groups: inverse reinforcement learning-based methods and imitation learning-based methods. 
For methods grounded in inverse reinforcement learning, certain studies in general-purpose planning \cite{rosbach2019driving,rosbach2020driving} explore style learning within complex urban environments. Conversely, other researchers utilize inverse reinforcement learning to acquire the linear-structure cost function, thereby enabling user-style autonomous driving \cite{kuderer2015learning, wu2020efficient, huang2021driving}. These methodologies learn user-style cost functions from demonstrations to achieve personalized planning. Besides, 
the methods based on imitation learning \cite{vitelli2022safetynet, pini2023safe, hu2023planning, huang2023conditional, fu2023interactionnet, hu2024pre} often aim to obtrain a human-like driving strategy by learning from large-scale human expert demonstrations.
The above methods have demonstrated promising results in their original tasks. However, when it comes to personalized planning, user demonstrations are expensive and limited in quantity. In the experiment (refer to Table. \ref{table:open_loop_test}), we find that fine-tuning the pre-trained model using the above methods on sparse user demonstrations resulted in degraded planning performance and poor generalization of driving styles.

To address the above challenges, we propose an effective transfer imitation learning method based on the four design principles proposed in gapBoost \cite{wang2019transfer}. Here's how we integrate these principles: \textbf{Rule 1}: We construct a weighted experience loss based on the imitation learning method. \textbf{Rule 2}: We assign equal weights to the samples in the same domain to avoid highly expensive computations. \textbf{Rule 3}: We adjust the sampling probabilities of source and target datasets by modifying the ratio of expert samples to user samples in the input batch. Since the amount of expert data vastly exceeds the amount of user data, there is a higher expectation value for the weights assigned to user samples. \textbf{Rule 4}: We utilize the inverse reinforcement learning method to construct the regularization term, which measures the style gap by calculating the style feature expectation error of the trajectory. We implicitly learn the weight of each sample by adjusting the model parameters. 

We conduct numerous experiments on the proposed method. We apply the proposed method to evaluate the performance of various fine-tuning structures. Our method outperforms all baseline methods in both style error and planning performance. It is worth noting that the baseline approaches achieved worse results than the pre-trained model, while our approach achieved superior performance. Additionally, through the sufficient update steps, we find that the fine-tuning architecture integrating a neural network with a differentiable nonlinear optimizer yields the best performance.

In summary, this paper makes the following three contributions:
\begin{itemize}
    \item In this paper, we propose a novel approach called the instance-based transfer imitation learning method for achieving personalized planning within intricate urban settings.
    \item The proposed method effectively transfers knowledge from expert data, addressing the challenges of overfitting and poor generalization stemming from the sparse user demonstrations.
    \item The effectiveness of the proposed method is demonstrated through a multitude of experiments. Experimental results indicate that our method achieves superior user style approximation and planning performance enhancement.
\end{itemize}

\section{Related Work}
% In this section, we first discuss machine learning-based planning, followed by personalized motion planning for autonomous driving.
% 1. 基于机器学习的轨迹规划(端到端)
\subsection{Machine Learning-based Planning for Autonomous Driving}
In the urban structured scene, some traditional trajectory planning methods based on optimization have been widely used in industry and academia \cite{sun2022fiss,fan2018baidu,ajanovic2018search,karaman2011sampling}. However, these hand-designed methods tend to have weak decision-making capabilities, cannot handle long-tail scenarios, and do not improve with increasing data. In recent years, more and more works have proved the potential of planning methods based on machine learning in handling a large number of different autonomous driving urban scenarios \cite{vitelli2022safetynet,pini2023safe,phan2023driveirl,hu2023planning,huang2023conditional,fu2023interactionnet,hu2024pre}. But machine learning-based planning often has difficulty interpreting its output, and when it encounters out-of-distribution situations it is likely to lead to dangerous actions.

Some hybrid frameworks combine the machine learning-based approach and optimization-based approach \cite{pulver2021pilot, huang2023differentiable}. Although it nicely combines the strengths of the machine learning-based approach and the optimization-based approach, it requires large amounts of human data to train.

In the case of sparse user demonstrations, it is difficult for the previous method to obtain a sufficiently safe policy while learning styles, and there are problems of overfitting and poor generalization. To address these issues, this paper proposed personalized planning via transfer imitation learning, which is based on pre-training and fine-tuning framework. It can learn style from user data and improve the performance of planning.

% 2.个性化的驾驶风格(基于偏好的规划) 个性化风格学习
\subsection{Personalized Planning for Autonomous Driving}
Personalized planning stands as a crucial element in enhancing the user experience. Some optimization-based approaches \cite{yan2022driver,zhang2022personalized,oudainia2022personalized} tend to focus solely on specific scenarios or tasks such as lane changes and obstacle avoidance. Moreover, certain studies on user style learning \cite{kuderer2015learning, wu2020efficient, huang2021driving} frequently employ the Maximum Entropy Inverse Reinforcement Learning (MaxEnt-IRL) \cite{ziebart2008maximum} method to learn a cost function for modeling the user's driving style. However, applying this method to complex urban scenes poses significant challenges. Furthermore, while the general-purpose planning approach considers style learning within complex urban scenarios \cite{rosbach2019driving,rosbach2020driving}, it cannot utilize data for continuous enhancement of planning performance.

It often proves challenging for these endeavors to concurrently accomplish vital aspects of both planning tasks. These include personalized planning for complex urban scenarios and the ongoing enhancement of planning performance through data utilization. This work addresses the issues of sparse user data and data-driven planning by transferring knowledge from expert data.

\subsection{Instance-Based Transfer Learning}
In the context of personalized planning tasks in autonomous driving scenarios, the data from the expert domain and user domain exhibit strong similarities. In this case, the instance-based transfer learning approach emerges as a promising solution.
Several case-based transfer learning methods employ the calculation of sample weights to facilitate effective sample transfer \cite{pardoe2010boosting,cai2019probabilistic,cai2019probabilistic,gupta2023boosting,jiang2019deep,wang2019transfer}.
Furthermore, some endeavors concentrate on screening valid samples from either the target or source domain \cite{wang2019instance,tang2024selecting}.

Inspired by these methods, we devised an effective transfer imitation learning method. Through data classification remove invalid entries. Additionally, our method implicitly adjusts sample weights via sample sampling probability and model parameter learning, thus circumventing the computational overhead associated with explicit sample weight calculation.

\begin{figure*}[t]
    \vspace{0.2cm}
    \centering
    \includegraphics[width=2.0\columnwidth]{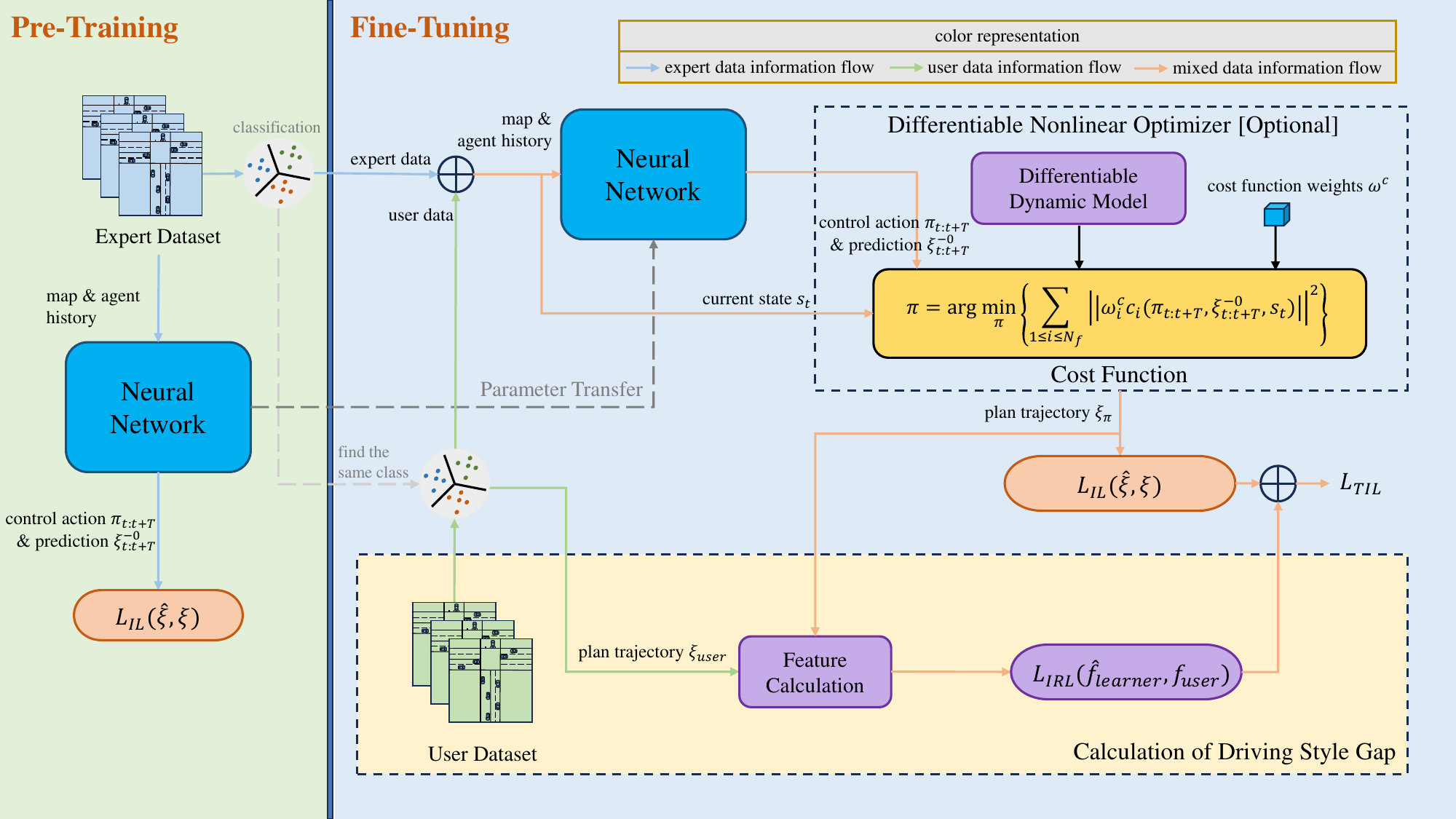}
    \vspace{-0.3cm}
    \caption{Personalized planning framework based on transfer imitation learning. For the specific structural design of the neural network module shown in the figure, we refer to DIPP \cite{huang2023differentiable}. During the pre-training stage, we train the neural network using extensive expert data. In the fine-tuning phase, we initialize the neural network with the pre-trained parameters and employ a combination of expert and user data as input batches to the model. Additionally, the differentiable motion planner module can be optionally utilized at the output side of the neural network for end-to-end fine-tuning. We leverage the maximum entropy inverse reinforcement learning method to construct $\mathcal{L}_{IRL}$. This term aiming to match the user trajectory feature expectation and learn the user trajectory style. And $\mathcal{L}_{IL}$ serves to minimize the experience error of trajectories, thereby ensuring the effectiveness of the planning.}
    \vspace{-0.1cm}
    \label{fig:system_overview_figure}
\end{figure*}
\section{Personalized Planning via Transfer Imitation Learning}

\subsection{Problem Formulation}
% In this work, the driving scene is abstracted as a continuous-space discrete-time system. 
We assume the current time step is $t$. The agent state at the current time is represented as $\bm{s}_t=[x_t,y_t,\theta_t,v_t]$, where $(x_t,y_t)$ is two-dimensional space coordinate, $\theta_t$ is heading angle, and $v_t$ is velocity. The trajectory $\xi_{0:t}$ is represented as a set of discrete points $\{s_0,...,s_t\}$. The model predicts the future trajectory of agent $i$ for the next $T$ time steps, and it's represented as $\hat{\xi}^i_{t:t+T}$. The output of the neural network is the control action sequence and denoted as ${\Pi}^m_{t:t+T}(m={1,..., N_m})$, 
where $N_m$ is the number of multi-modal combinations. Specifically, the selected optimal control action sequence is denoted as ${\pi}_{t: T}$. All learnable parameters are denoted as $\omega$, including the neural network parameters $\omega^{n}$ and the cost function weights $\omega^{c}$.
% The cost function weights are denoted as $\omega_i(i={1,..., N_f})$. %, where $N_f$ is the number of cost functions $c_i(u)$.

\subsection{Pre-Training with Large-Scale Expert Data} % Imitation Learning
% 运动学模型 
\subsubsection{Differentiable Kinematic Model} To obtain a planning trajectory that satisfies the kinematic constraints, the neural network outputs the control action $u_t=\{a_t,\delta_t\}$ (where $a_t$ is acceleration and $\delta_t$ is steering angle) and the trajectory is calculated through the kinematic model:
\begin{align}
\pi_{t:t+T} &= \begin{bmatrix} u_t & \dots & u_T \end{bmatrix}^\intercal\\
\hat{\xi}^0_{t:t+T} &= \psi(s_t, \pi_{t:t+T})
\end{align}
where the $\psi$ represents the kinematic model. In this work, we adopt the differentiable kinematic bicycle model in \cite{huang2023differentiable}.

% The details of the calculation processes are formulated as follows: 
% \begin{equation}
% \begin{aligned}
% \begin{bmatrix} x_{t+1} \\ y_{t+1} \\ \theta_{t+1} \\ v_{t+1} \end{bmatrix}
% = 
% \begin{bmatrix} 
% 1 & 0 & 0 & \cos(\theta_t)\Delta t \\
% 0 & 1 & 0 & \sin(\theta_t)\Delta t \\
% 0 & 0 & 1 & \frac{1}{C_L}\tan(\delta_t)\Delta t \\
% 0 & 0 & 0 & 1 \\
% \end{bmatrix}
% \begin{bmatrix} 
% x_{t} \\ y_{t} \\ \theta_{t} \\ v_{t}
% \end{bmatrix}
% +
% \begin{bmatrix} 
% 0 \\ 0 \\ 0 \\ a_t\Delta t
% \end{bmatrix}
% \end{aligned}\label{eq:kinematic_model}
% \end{equation} 
% where $\Delta t$ is the time interval and $C_L$ is the wheelbase of the vehicle. 

% \begin{equation}
% \begin{aligned}
% &x_{t+1} = x_{t} + v_t\cos(\theta_t)\Delta t\\  
% &y_{t+1} = y_{t} + v_t\sin(\theta_t)\Delta t\\  
% &\theta_{t+1} = \theta_{t} + \frac{v_t}{C_L}\tan(\delta_t)\Delta t\\  
% &v_{t+1} = v_{t} + a_t\Delta t\\
% \end{aligned}\label{eq:kinematic_model}
% \end{equation}

\subsubsection{Neural Network}
In the pre-training stage of imitation learning, we adopt the same structure as DIPP \cite{huang2023differentiable}. The pre-training loss is summarized as follows: 
\begin{equation}
\begin{aligned}
\mathcal{L}_{IL} = \lambda_1\mathcal{L}_{prediction} +\lambda_2\mathcal{L}_{score} +\lambda_3\mathcal{L}_{imitation}  
\end{aligned}\label{eq:il_loss}
\end{equation}
where $\lambda_i(i=1,2,3)$ is the weight that scales the different loss terms. 

For imitation loss, the control action sequence is used to calculate the trajectory of the autonomous vehicle through the kinematic model, and the distance from the ground truth is measured to form the imitation loss: 
% 公式:Imitation loss
\begin{equation}
\begin{aligned}
\mathcal{L}_{imitation} = \mathrm{smooth} L1(\psi(s_t, \pi_{t:t+T}) - \xi^0_{t:t+T})
\end{aligned}
\end{equation}
% 其中$\xi^0_{t:t+T}$表示数据集中的真实轨迹。
where \(\xi^0_{t:t+T}\) represents the ground truth trajectories in the dataset.

\subsection{Fine-Tuning with Instance-Based Transfer Imitation Learning}
\subsubsection{Differentiable Nonlinear Optimization} 
In the fine-tuning stage, we use differentiable nonlinear optimization \cite{pineda2022theseus} to refine the control action sequence of the output of the pre-training model.
To be specific, the optimization process is formulated as follows:
% modified N_f>2
\begin{equation}
\begin{aligned}
{\pi}^* = \arg \min_{\pi}\ \frac{1}{2}\sum_i^{N_f}\Vert \omega_i^c c_i(\pi)\Vert^2
\end{aligned}\label{eq:second_stage_obj}
\end{equation}
where $c_i(\pi)$ is the cost function of the control action sequence $\pi$, $\omega_i^c(i=1,...,N_f)$ is the scaling weight and $N_f$ is the number of cost functions $c_i(\pi)$. The trajectory features are designed based on autonomous driving planning tasks, which include travel efficiency, ride comfort, lane departure, traffic rules, and most importantly safety \cite{huang2023differentiable}.

% 最大熵IRL
\subsubsection{Maximum Entropy Inverse Reinforcement Learning}.
To solve the ambiguity of multiple solutions and the randomness of expert behavior, we adopt Maximum Entropy Inverse Reinforcement Learning (MaxEnt-IRL) to learn the weights \cite{ziebart2008maximum}:
\begin{equation}
\begin{aligned}
P(\xi|\omega^c) = \frac{\exp(-c(\xi|\omega^c))}{Z(\omega^c)}
\end{aligned}\label{eq:second_stage_obj}
\end{equation}
where $Z(\omega^c)$, called partition function, equals to $\sum_\xi\exp(-c(\xi|\omega^c))$. According to this function, plans with equivalent costs have equal probability.

The Max-Ent IRL method optimizes the weight $\omega^c$ by maximizing the probability of the human demonstration trajectories $\xi \in D$. The gradient of this optimization problem can be computed as follows \cite{ziebart2008maximum}:

% It is a Maximum Likelihood Estimation (MLE) process and the optimization objective is as follows: 
\begin{equation}
\begin{aligned}
\nabla \mathcal{L}_{IRL} = \mathbb{E}_{P(\xi|\omega^c)}[\hat f_i]-f_i
\end{aligned}\label{eq:nabla_irl_loss}
\end{equation}

However, calculating the expectation in Eq. \ref{eq:nabla_irl_loss} through sampling is challenging. One possible approximation of the expected feature values is to compute the feature values of the most likely trajectory \cite{kuderer2015learning}. To extend to neural network parameter updates, instead of using feature errors as gradients, we approximate the solution by performing gradient descent on the following objective: 

\begin{equation}
\begin{aligned}
\mathcal{L}_{IRL} \approx \frac{1}{N_f} \sum_{i=1}^{N_f}{\beta_i|\hat f_i-f_i|}
\end{aligned}\label{eq:irl_loss}
\end{equation}
where \(f\) is the feature of the trajectory and \( \beta \) is the corresponding scaling factor. We refer to \cite{huang2023differentiable} for the design of seven metrics to describe trajectory features. The name and scaling factor of each feature are shown in Table \ref{table:irl_scaling}.

\begin{table}[h] 
\vspace{-0.4cm}
\hspace{-0.1cm}
\renewcommand{\arraystretch}{1.5}
        \caption{The feature weights in inverse reinforcement learning} 
	\scalebox{0.88}{
	\setlength{\tabcolsep}{2.5pt}{
	\begin{tabular}{c|ccccccc} 
		\toprule[1pt] 
		features&Acc(lat)&Acc(lon)&Jerk(lat)&Jerk(lon)&Efficiency&Road Offset&Safety\\
		\cmidrule(lr){1-8}
        scaling&0.008&0.008&0.004&0.01&0.004&0.0005&0.01\\
		\bottomrule[1pt] 
	\end{tabular}
        \label{table:irl_scaling}
        } 
	}
    \begin{tablenotes}
    \footnotesize
    \item Acc(lat): lateral acceleration, Acc(lon): longitudinal acceleration.
    \end{tablenotes}
 \vspace{-0.1cm}
\end{table}

\subsubsection{Instance-Based Transfer Imitation Learning}
By adhering to the design principles delineated in gapBoost \cite{wang2019transfer}, we propose an instance-based transfer imitation learning algorithm. By amalgamating Eq. \ref{eq:irl_loss} and Eq. \ref{eq:il_loss}, we formulate the optimization objectives of instance-based transfer imitation learning as follows:
% By referring to the design principles proposed in gapBoost[cite,gapBoost], we propose an Instance-Based Transfer Imitation Learning algorithm. Specifically, we set the weights of equal samples to avoid the expensive calculation cost of weight calculation. At the same time, we randomly sample expert data and user data from the data set in a fixed proportion to form the input batch data $D^m$. Since the user data in the target domain is much less than the source data, the target domain sample will get a higher weight expectation. Finally, by combining Eq. \ref{eq:irl_loss} and Eq. \ref{eq:il_loss}, we obtain the optimization objectives of Instance-Based Transfer Imitation Learning as follows:
\begin{equation}
\begin{aligned}
\mathcal{L}_{TIL}^\alpha = \mathcal{L}_{IL} + \alpha\mathcal{L}_{IRL}
\end{aligned}\label{eq:til_loss}
\end{equation}
The parameter $\alpha > 0$ denotes the scaling factor of the $L_{IRL}$ term, utilized for adjusting the weight of the regularization component within the optimization objective. $L_{IRL}$ serves as a regularization term aimed at reducing the performance disparity across domains, which accomplishes style approximation by aligning the feature distribution expectation of the input sample with that of the user sample. 
In practical implementation, we utilize this regularization term along with an imitation learning term to update all learnable parameters $\omega$.

\begin{table}[h] 
\vspace{-0.1cm}
\renewcommand{\arraystretch}{1.50}
        \caption{The classification rules for planning trajectory} 
	\scalebox{0.88}{
	\setlength{\tabcolsep}{2.5pt}{
	\begin{tabular}{c|c|c} 
		\toprule[1pt] 
		classification rule&rule1&rule2\\
		\cmidrule(lr){1-3}
            Stationary&$max(speed) < 2 m/s$ &$Displacement<2 m$\\
            \cmidrule(lr){1-3}
            Straight&$abs(\delta_{heading}) < \frac{\pi}{6}$&--\\
            \cmidrule(lr){1-3}
            \multirow{2}{*}{Turn}&$abs(\delta_{heading}) < -\frac{\pi}{6}$&$abs(\delta_{heading}) > \frac{\pi}{6}$\\
            &$Lateral Displacement>0$&$Lateral Displacement<0$\\
		\bottomrule[1pt] 
	\end{tabular}
        \label{table:classification}
        } 
	}
    % \begin{tablenotes}
    % \footnotesize
    % \item Acc: Acceleration, Acc(lat): Lateral Acceleration
    % \end{tablenotes}
 \vspace{-0.2cm}
\end{table}

\begin{algorithm}[ht!]
\setlength{\intextsep}{10pt plus 1pt minus 1pt}
\caption{Personalized planning via Transfer Imitation Learning (PP-TIL)}\label{algo:PP-TIL}
% \SetAlgoLined
% \SetKwInOut{Input}{Input}\SetKwInOut{Output}{Output}
% \Input{Human demonstration trajectory dataset $\mathcal{D}=\{ \zeta_i \}_{i=1}^N$, environment model $P$, learning rate $\alpha$, regularization parameter $\lambda$, number of epochs $E$}
% \Output{Optimized reward function parameters $\boldsymbol \theta^*$}
% \BlankLine

\textbf{notation}: model parameters $\omega$, class $k$ expert dataset $\mathcal{D}_{expert}^k$, class $k$ user dataset $\mathcal{D}_{user}^k$, class $k$ expert samples $X_{expert}^k$, class $k$ user samples $X_{user}^k$, class $k$ input batch $X_{input}^k$, expert proportion in a batch $\mathcal{P}$, batch size $\mathcal{N}$, update step $\mathcal{S}$, class number $K$.\\
\textbf{initialize}: $\omega^{n} \leftarrow $ initialize parameter via PreTraining (Eq. \ref{eq:il_loss})\\
\textbf{[optional]} joint differentiable nonlinear optimizer\\
\textbf{Classification}: a rule-based approach is used to classify the dataset into $K$ classes

\For{$step \in \{1, ..., \mathcal{S}\}$}{
    \For{$k \in \{1, ..., K\}$}{
    $X_{expert}^k \leftarrow$ sample $\mathcal{N}*\mathcal{P}$ data frames from $\mathcal{D}_{expert}^k$
    
    $X_{user}^k \leftarrow$ sample $\mathcal{N}*(1-\mathcal{P})$ data frames from $\mathcal{D}_{user}^k$
    
    $X_{input}^k \leftarrow$ concatenate $X_{expert}^k$ and $X_{user}^k$

    % (X_{input}^k;\mathcal{D}_{user}^k)
    $\mathcal{L}_{IRL} \leftarrow$ build IRL loss based on $X_{input}^k$ and $\mathcal{D}_{user}^k$
    
    $\omega^* \leftarrow$ update via $X_{input}^k$ and Eq. \ref{eq:til_loss}
    }
}
\textbf{return} $\omega^*$ % $ \leftarrow \omega$
\end{algorithm}
As shown in Algo. \ref{algo:PP-TIL}, We delineate the detailed process of instance-based transfer imitation learning. Initially, we categorize the dataset based on the type of planning trajectory. Refer to Table \ref{table:classification} for the specific classification criteria. In our study, we categorize it into three classes: stationary, straight, and turn. As the stationary category lacks user style information, we exclude it during the fine-tuning phase. We sample from both the expert and user data within the remaining categories to construct the mixed input batch at each iteration. Simultaneously, we utilize the corresponding user data category to formulate the IRL regularization term. Finally, we feed the batch data into the model for gradient computation and update the parameters using $\mathcal{L}_{TIL}$.
% We show the specific process of instance-based transfer imitation learning in detail in Algo. \ref{algo:PP-TIL}. First, we classify the data set by type. In this work, we divide it into three categories according to the type of planned trajectory: Stop, go straight, bend. Since the Stationary category data does not carry any user style information, we removed this category during the fine-tuning process. The specific classification criteria are shown in Table \ref{table:classification}. After filtering the invalid categories, we take a proportional number of samples from the expert data and user data of the remaining categories at each step to form the input batch. At the same time, we use the corresponding category of user data to establish IRL optimization goals. Finally, we enter batch data to gradient the model and update the parameters with $\mathcal{L}_{TIL}$.

\section{Experiments}
\subsection{Experiment Settings}
\subsubsection{Dataset Settings}
We train and validate the approaches on the Waymo Open Motion Dataset (WOMD) \cite{ettinger2021large}, a large-scale real-world driving dataset focus on urban driving scenarios. 
We set a 7-second time window to segment the 20-second driving scene into frames, which includes a 2-second horizon for historical observation and a 5-second horizon for future prediction and planning. The window slides in 10 timesteps from the beginning of the scene and produces segments of 14 frames. In the experiment, we obtain 916808 frames from the scene, of which 80\% is used for training, while the rest is used for open-loop testing. 
% Furthermore, the trajectory feature vector is computed based on the feature terms and scaling coefficients outlined in Table \ref{table:irl_scaling}. Subsequently, 10,000 samples are extracted from the test set for clustering using K-means, resulting in three clustering centers. Finally, we select the 64 sample frames nearest to each center point to obtain three distinct user datasets representing different styles. 

In addition, based on the feature items and scaling factors in Table \ref{table:irl_scaling}, we compute trajectory feature vectors. Subsequently, we extract 10,000 samples from the test set and apply the k-means clustering algorithm to obtain three clustering centers. Finally, we select the 64 sample frames closest to each cluster center to form three distinct datasets, which we use as user data for training. The expectation of user style features is presented in Table \ref{table:user_style}.

\begin{table}[h] 
% \hspace{0.1cm}
\renewcommand{\arraystretch}{1.7}
        \caption{The style feature expectations of the different users}
	\scalebox{0.80}{
	\setlength{\tabcolsep}{2.5pt}{
	\begin{tabular}{c|c|ccccccc} 
		\toprule[1pt] 
		class&name&Acc(lat)&Acc(lon)&Jerk(lat)&Jerk(lon)&Efficiency&Road Offset&Safety\\
        \cmidrule(lr){1-9}
        \multirow{3}*{\rotatebox{90}{\makecell[c]
        {Straight}}}
        &User1&0.03089&0.00030&0.00620&0.00417&0.03358&
        0.00671&0.00127\\
        &User2&0.10873&0.00549&0.00001&0.00166&0.04181&
        0.00465&0.00073\\
        &User3&0.00126&0.00030&0.00333&0.00123&0.02078&
        0.00678&0.00002\\
        \cmidrule(lr){1-9}
        \multirow{3}*{\rotatebox{90}{\makecell[c]{Turn}}}
        &User1&0.04955&0.16039&0.05637&0.01409&0.03583&
        0.02098&0.00207\\
        &User2&0.26095&0.11560&0.02183&0.01154&0.04054&
        0.02136&0.00141\\
        &User3&0.00807&0.01139&0.03083&0.01343&0.03805&
        0.01948&0.00305\\
		\bottomrule[1pt] 
	\end{tabular}
        \label{table:user_style}
        } 
	}
    \begin{tablenotes}
    \footnotesize
    \item Acc(lat): lateral acceleration, Acc(lon): longitudinal acceleration.
    \end{tablenotes}
 % \vspace{-0.5cm}
\end{table}

% \subsection{Metrics}
\subsubsection{Metrics}
In the experiment, this paper designs collision rate, red light, off route, planning error, and prediction error referring to DIPP \cite{huang2023differentiable}. 
% To evaluate the effect of the algorithm more completely, the over speed metric is designed in this paper. We consider the output trajectory of the algorithm exceeding the road limit speed as over speed and count the total number of frames violating the rule. 
Additionally, we define the ``style error" metric, calculated as shown in Eq. \ref{eq:irl_loss}. The ``style error" is evaluated by computing the mean absolute error between the trajectory features produced by the model and those from the user demonstration, thus assessing the style similarity.
\subsubsection{Implementation Details}
The pre-trained model undergoes five epochs of training on the expert dataset. Throughout training, the step size of the nonlinear optimizer is fixed at 0.4, and the number of iterations is set to 2. All experiments in this paper adhere to the same initial parameter settings as follows: a batch size of 64, a neural network learning rate of 1e-5, and a linear weight learning rate for the cost function of 1e-3.

\subsection{Open-Loop Test}
\begin{table}[h]
\vspace{0.2cm}
% \vspace{-0.1cm}
\renewcommand{\arraystretch}{1.7}
        \caption{Hand-Crafted cost function weights} 
	\scalebox{1.0}{
	\setlength{\tabcolsep}{2.5pt}{
	\begin{tabular}{c|ccccccccc}
		\toprule[1pt] 
		cost function&speed&acc&jerk&steer&rate&pos& head&traffic&safety\\
		\cmidrule(lr){1-10}
        weights&0.1&0.5&0.1&0.01&0.5&0.5&5&10&10\\
		\bottomrule[1pt] 
	\end{tabular}
        % \label{table:human_weights}
        } 
	}
    \begin{tablenotes}
    \footnotesize
    \item The specific details of the cost function can be found in DIPP \cite{huang2023differentiable}.
    \end{tablenotes}
 \vspace{-0.4cm}
 \label{table:cost_function_scaling}
\end{table}

\begin{table*}[htp!]
\vspace{0.2cm}
\renewcommand{\arraystretch}{1.55}
\caption{Compare fine-tuning methods in open-loop testing} 
% \scalebox{1.0}{
% \renewcommand{\arraystretch}{1.25}
\setlength{\tabcolsep}{4.0pt}{
% \centering
\begin{threeparttable}
\begin{tabular}{@{}|c|c|cc|c|ccc|cc|c|@{}}
\toprule
\multirow{3}{*}{\makecell[c]{Fine-Tuning\\Structure}} &\multirow{3}{*}{\makecell[c]{Parameter Update\\(Updated Part / Method)}}&
\multicolumn{3}{c|}{In Domain}&\multicolumn{6}{c|}{Out of Domain}\\
&&
\multicolumn{2}{c|}{Plan error ($m$)} &
\multicolumn{1}{c|}{Style} &
% \multicolumn{1}{c}{Over speed} &
\multicolumn{1}{c}{Red light} &
\multicolumn{1}{c}{Off route} &
\multicolumn{1}{c|}{Collision} &
\multicolumn{2}{c|}{Prediction error ($m$)} &
\multicolumn{1}{c|}{Style} \\
&& @1s& @3s&error& (\%)$\downarrow$&  (\%)$\downarrow$&(\%)$\downarrow$& ADE@5s&FDE@5s&error\\
\hline
\hline
\multirow{4}{*}{NN}
&None& 0.2044& 0.5994& 0.08321& 0.12& 0.42& 3.14& 0.6857& 1.7015& 0.06892\\
\cline{2-11}
&NN / $\mathcal{L}_{IL}$& \textbf{0.1991}& \textbf{0.4637}& 0.06793& 0.78& \textbf{0.42}& 3.80& 0.7086& 1.7273& 0.06545\\
&NN / $\mathcal{L}_{IRL}$& 0.2124& 0.7064& \textbf{0.02050}& 0.22& \textbf{0.42}& 3.86& 0.7000& 1.7154& 0.05255\\
&\cellcolor{lightgray}NN / $\mathcal{L}_{TIL}$ (Ours)&\cellcolor{lightgray} 0.2038&\cellcolor{lightgray} 0.5076&\cellcolor{lightgray} 0.03950&\cellcolor{lightgray} \textbf{0.19}&\cellcolor{lightgray} \textbf{0.42}&\cellcolor{lightgray} \textbf{2.87}&\cellcolor{lightgray} \textbf{0.6753}&\cellcolor{lightgray} \textbf{1.6579}&\cellcolor{lightgray} \textbf{0.04448}\\ 
\hline
\hline
\multirow{11}{*}{NN\&CF}
&None & 0.5861& 4.0462& 0.07570& 0.01& 0.27& 9.05& 0.6857& 1.7015& 0.05911\\
\cline{2-11}
&CF / Human& \textbf{0.2136}& 0.8657& 0.01748& 0.02& \textbf{0.39}& \textbf{1.94}& --& --& \textbf{0.03389}\\
&CF / $\mathcal{L}_{IL}$& 0.2162& \textbf{0.8579}& 0.03073& 0.02& 0.40& 3.90& --&--& 0.04329\\
&CF / $\mathcal{L}_{IRL}$ (LfD \cite{kuderer2015learning})& 0.4131& 1.4857& 0.04460& \textbf{0.01}& 0.42& 6.06& --&--& 0.04070\\
&\cellcolor{lightgray}CF / $\mathcal{L}_{TIL}$ (Ours)&\cellcolor{lightgray} 0.2725&\cellcolor{lightgray} 0.9494& \cellcolor{lightgray}\textbf{0.01695}&\cellcolor{lightgray}0.02&\cellcolor{lightgray} \textbf{0.39}&\cellcolor{lightgray} 2.82&\cellcolor{lightgray} --&\cellcolor{lightgray}--&\cellcolor{lightgray} 0.04247\\
\cline{2-11}
&NN / $\mathcal{L}_{IL}$ + CF / Human& 0.2284& 0.8787& 0.01724& \textbf{0.01}& \textbf{0.37}& 1.95& 0.7158& 1.7255& 0.03355\\
&NN / $\mathcal{L}_{IRL}$ + CF / Human& \textbf{0.2135}& 0.8843& \textbf{0.01522}&0.02& 0.38& 1.92& 0.7573& 1.7725& \textbf{0.03312}\\
&NN / $\mathcal{L}_{TIL}$ (Ours) + CF / Human& 0.2224& \textbf{0.8455}& 0.01675& 0.02& 0.38& \textbf{1.77}& \textbf{0.6865}& \textbf{1.6682}& 0.03342\\
\cline{2-11}
&NN\&CF / $\mathcal{L}_{IL}$ (DIPP \cite{huang2023differentiable})& \textbf{0.2458}& 0.8199& 0.02173& 0.03& \textbf{0.35}& 4.69& 0.7173& 1.7359& 0.03796\\
&NN\&CF / $\mathcal{L}_{IRL}$& 0.5738& 1.7309& 0.04342& \textbf{0.01}& 0.42& 8.69& 0.6919& 1.6929& 0.05499\\
& \cellcolor{lightgray}NN\&CF / $\mathcal{L}_{TIL}$ (Ours)&\cellcolor{lightgray} 0.2568&\cellcolor{lightgray} \textbf{0.7813}&\cellcolor{lightgray} \textbf{0.02063}&\cellcolor{lightgray} 0.02&\cellcolor{lightgray} 0.36&\cellcolor{lightgray} \textbf{3.58}&\cellcolor{lightgray} \textbf{0.6837}&\cellcolor{lightgray} \textbf{1.6756}& \cellcolor{lightgray}\textbf{0.03585}\\
\bottomrule
\end{tabular}
\begin{tablenotes}
\footnotesize
\item None: no fine-tuning, NN: neural network, CF: cost function, NN\&CF: joint neural network and cost function.
\end{tablenotes}
\end{threeparttable}
    }
% }
\label{table:open_loop_test}
\end{table*}

\begin{figure*}
   \centering
   \subfigure[Pre-Training Model]{\includegraphics[width=1.9in]{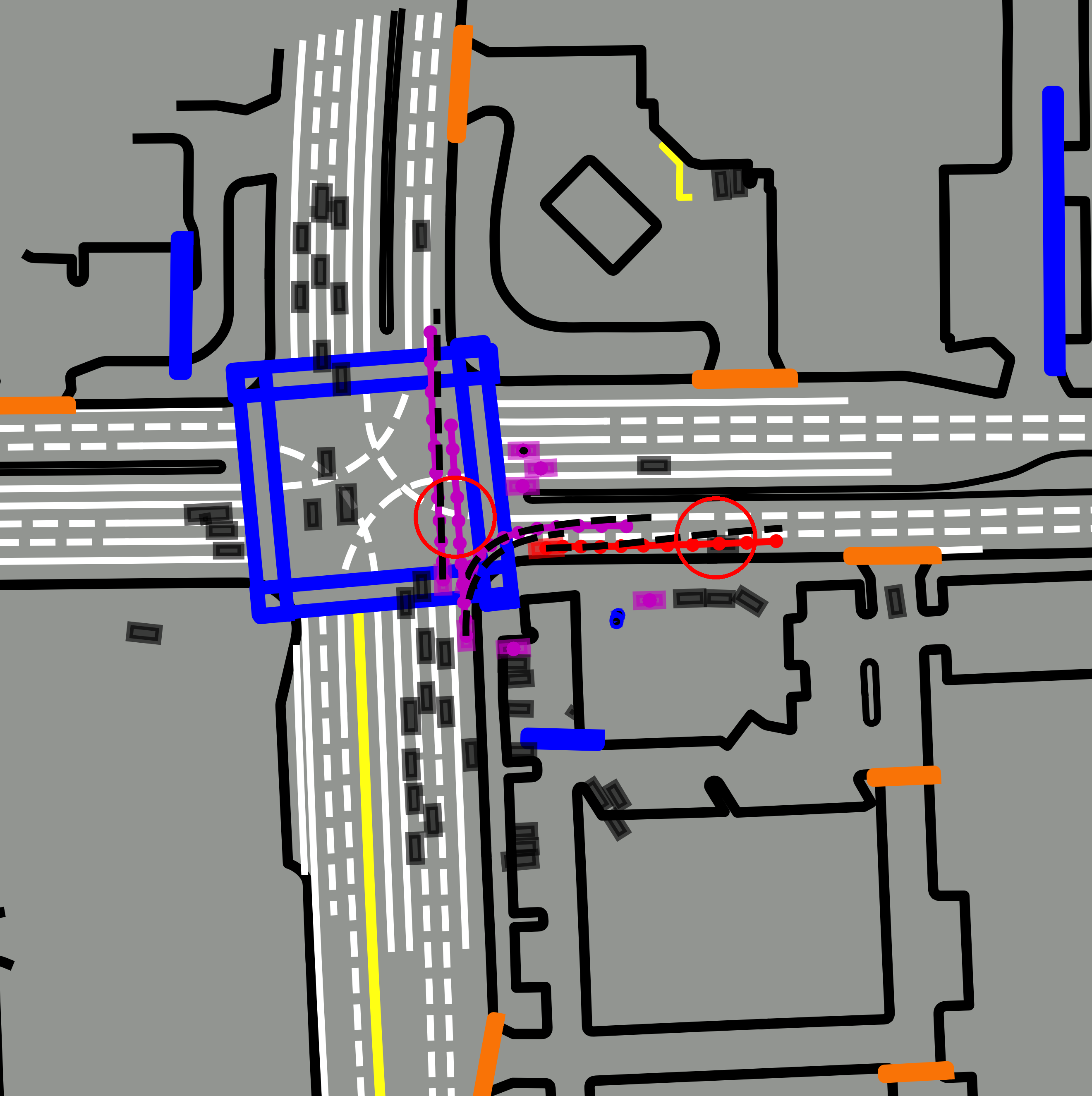}
    }% \label{fig:img0}
      % \hfill
   \hspace{1.0em}
   \subfigure[(NN) NN / Ours]{\includegraphics[width=1.9in]{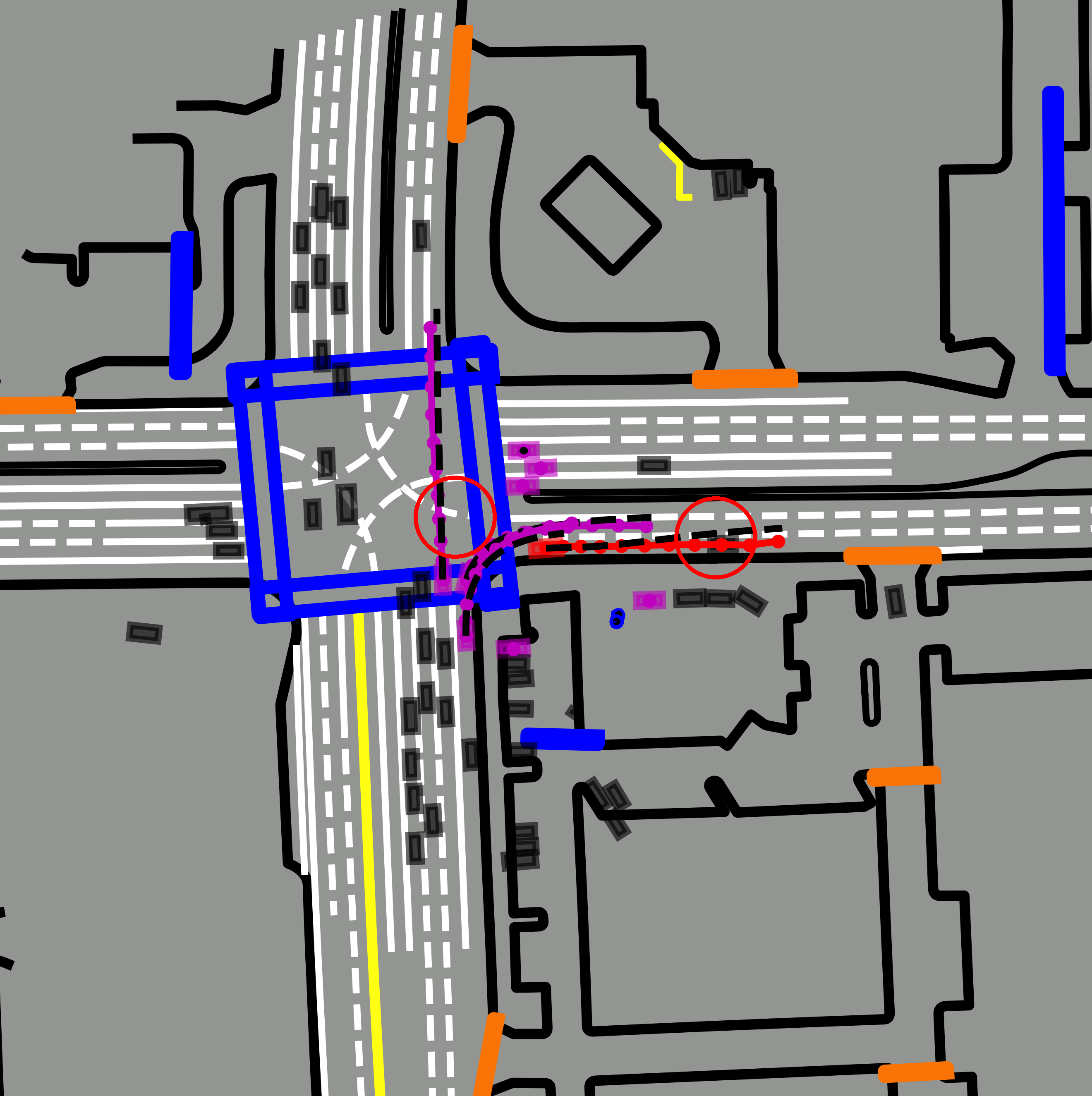}
      }% \label{fig:img1}
   % \hfill
   \hspace{1.0em}
   \subfigure[(NN\&CF) NN\&CF / Ours]{\includegraphics[width=1.9in]{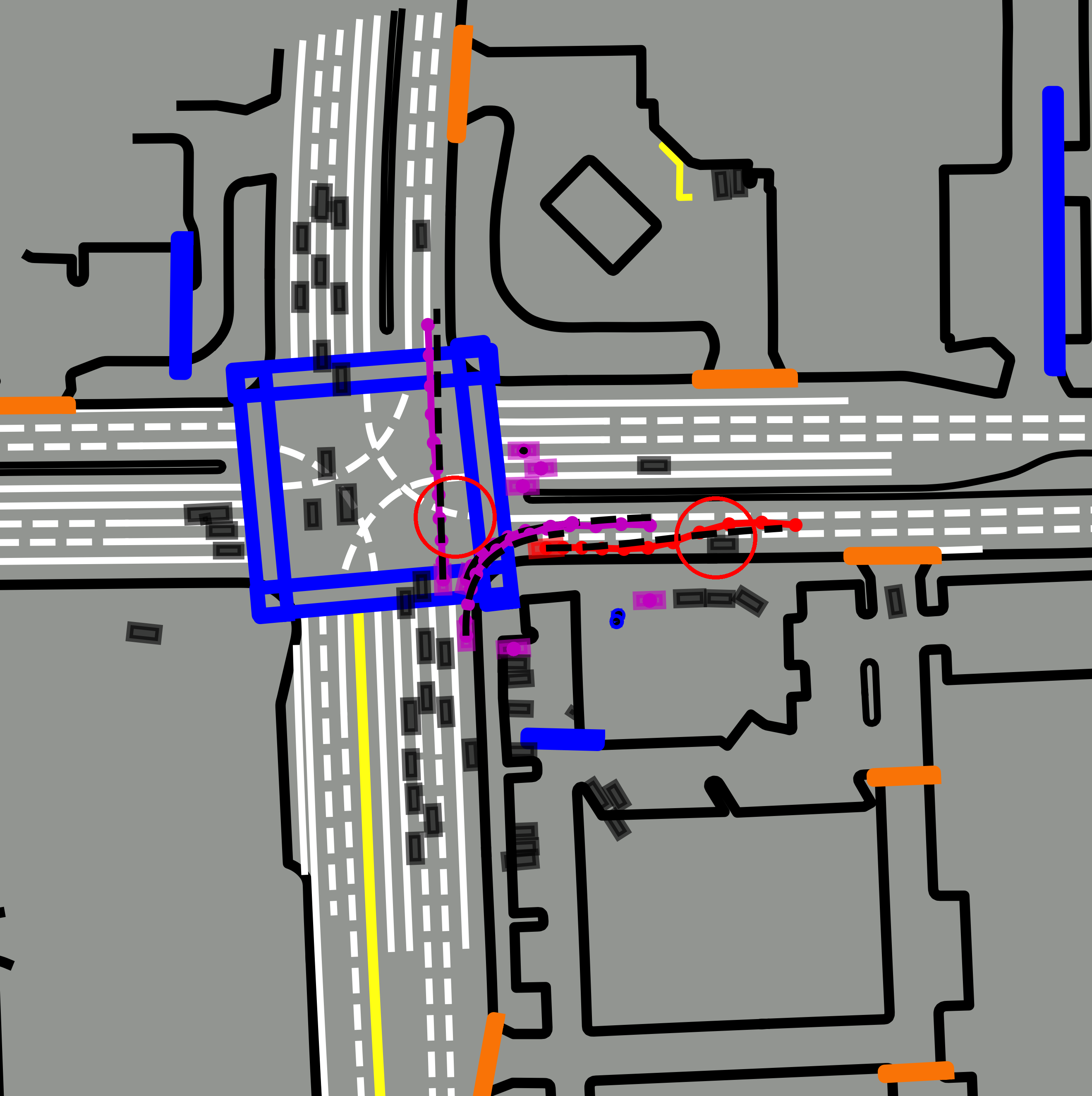}
      }\\ % \label{fig:img2}
   % \vspace{0.5em}

   % \includegraphics[width=6.6in]{figures/visual_compare_speed.png} % 6.6
   % \vspace{-0.2em}
   % \caption{Illustration of multi-style motion planning in urban autonomous driving scenarios. The red rectangle represents AV, and the red line represents the motion planning for the next 5 seconds under the current moment.}
   \caption{Visualization of the final results. We compare the user's real trajectory with the output trajectory of three different models. Different categories are visually distinguished using various colors in the diagram. The color scheme for the rectangles is depicted as follows: \textcolor[RGB]{255,0,0}{autonomous vehicle}, \textcolor[RGB]{255,0,255}{predicted vehicle}, \textcolor[RGB]{0, 0, 0}{other vehicle}, \textcolor[RGB]{0, 0, 255}{crosswalk} and \textcolor[RGB]{255,140,0}{speed bump}. The color scheme for the lines is depicted as follows: \textcolor[RGB]{255,0,0}{planned trajectory}, \textcolor[RGB]{255,0,255}{predicted trajectory} and \textcolor[RGB]{0, 0, 0}{road edges}. In particular, the black dotted line represents the \textcolor[RGB]{0, 0, 0}{ground-truth trajectory}. The red circle in the figure aids in better distinguishing the differences between different approaches.
   } 
   \label{fig:visual_compare}
   % \vspace{0.1cm}
\end{figure*}

As depicted in Table \ref{table:open_loop_test}, we conduct comprehensive experiments involving various fine-tuning architectures and fine-tuning methods. ``$\mathcal{L}_{IL}$" denotes the imitation learning method outlined in Eq. \ref{eq:il_loss}, while ``$\mathcal{L}_{IRL}$" represents the inverse reinforcement learning method illustrated in Eq. \ref{eq:irl_loss}. ``$\mathcal{L}_{TIL}$" signifies the utilization of the transfer imitation learning method detailed in Eq. \ref{eq:til_loss}. Due to the low effectiveness of the methods without the pre-training process, we establish a stronger baseline to demonstrate the superiority of our approach. Specifically, all experiments listed in the table are conducted based on the pre-trained model, and we compare the performance of different methods during the fine-tuning stage. We update each model 100 times and repeat each fine-tuning method thrice, presenting the average results in the table. ``Human" denotes manually adjusting the weights of the cost function, as specified in Table \ref{table:cost_function_scaling}. In this experiment, our method involves mixing expert data and user data in a 50\% ratio, with the $\alpha$ of the ``$\mathcal{L}_{TIL}$" set to 100. We perform the in-domain evaluation on the user dataset and the out-of-domain evaluation on the test dataset. 

The experimental results are presented in Table \ref{table:open_loop_test} demonstrate that the performance of baseline methods is worse after fine-tuning than before because of overfitting, while the proposed approach achieves fine-tuning performance improvements. In addition, our method effectively considers both user behavior imitation within the domain and style generalization outside the domain. Notably, our method achieves competitive planning performance, along with lower style error.

\begin{figure}[t]
    \vspace{0.1cm}
    \setlength{\abovecaptionskip}{0cm}
    % \vspace{-1.6cm}
    % \hspace{-0.2cm}
    % \subfigure{ % The style performance.
    \includegraphics[height=0.78\columnwidth]{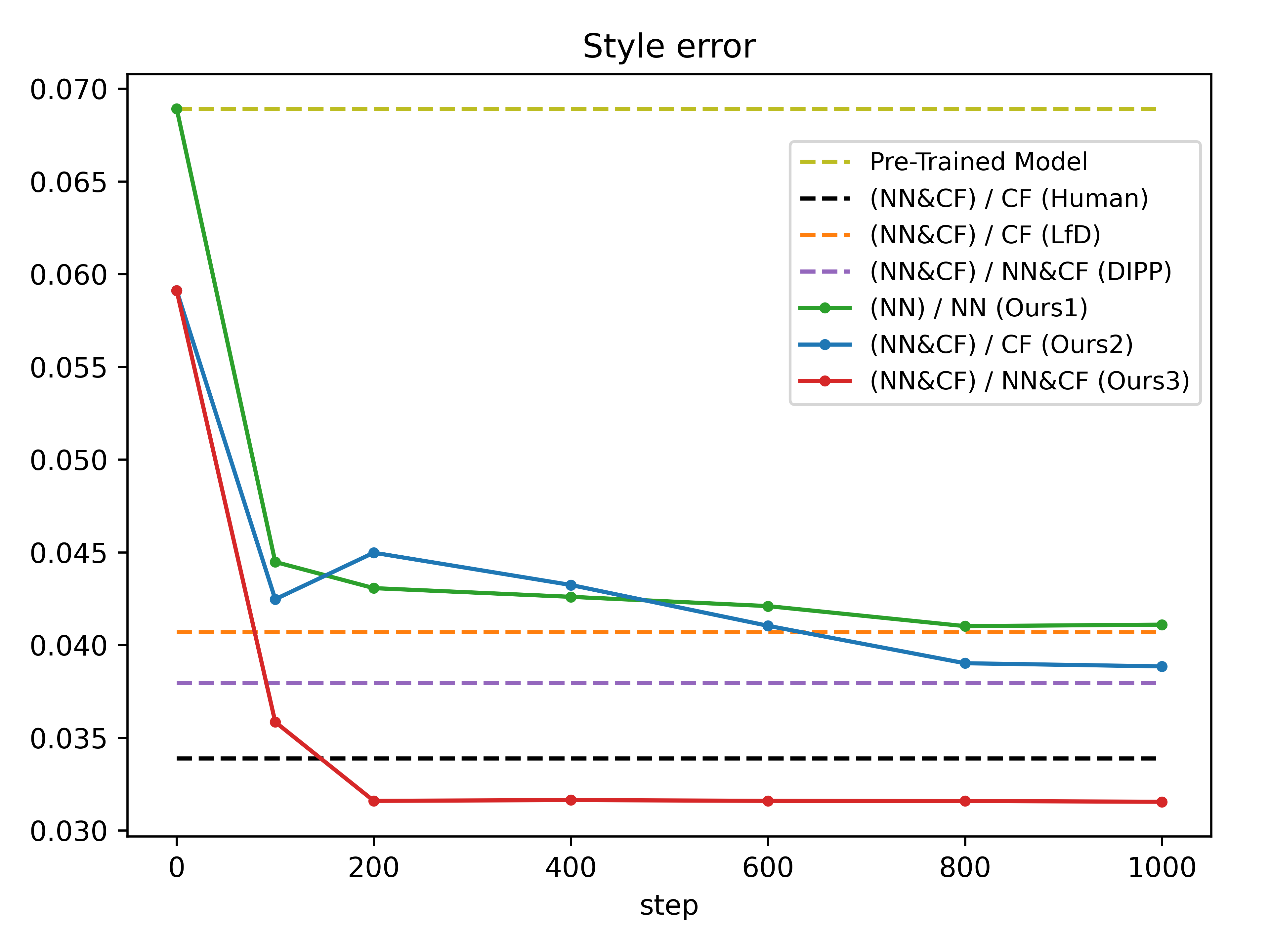}   
    % } % \label{feature_flow_local} 0.35
    % \hspace{-0.8cm}
    \vspace{-0.6cm}
    \\
    \hspace{0.3cm}
    % \subfigure{ % The safety performance.
    \includegraphics[height=0.75\columnwidth]{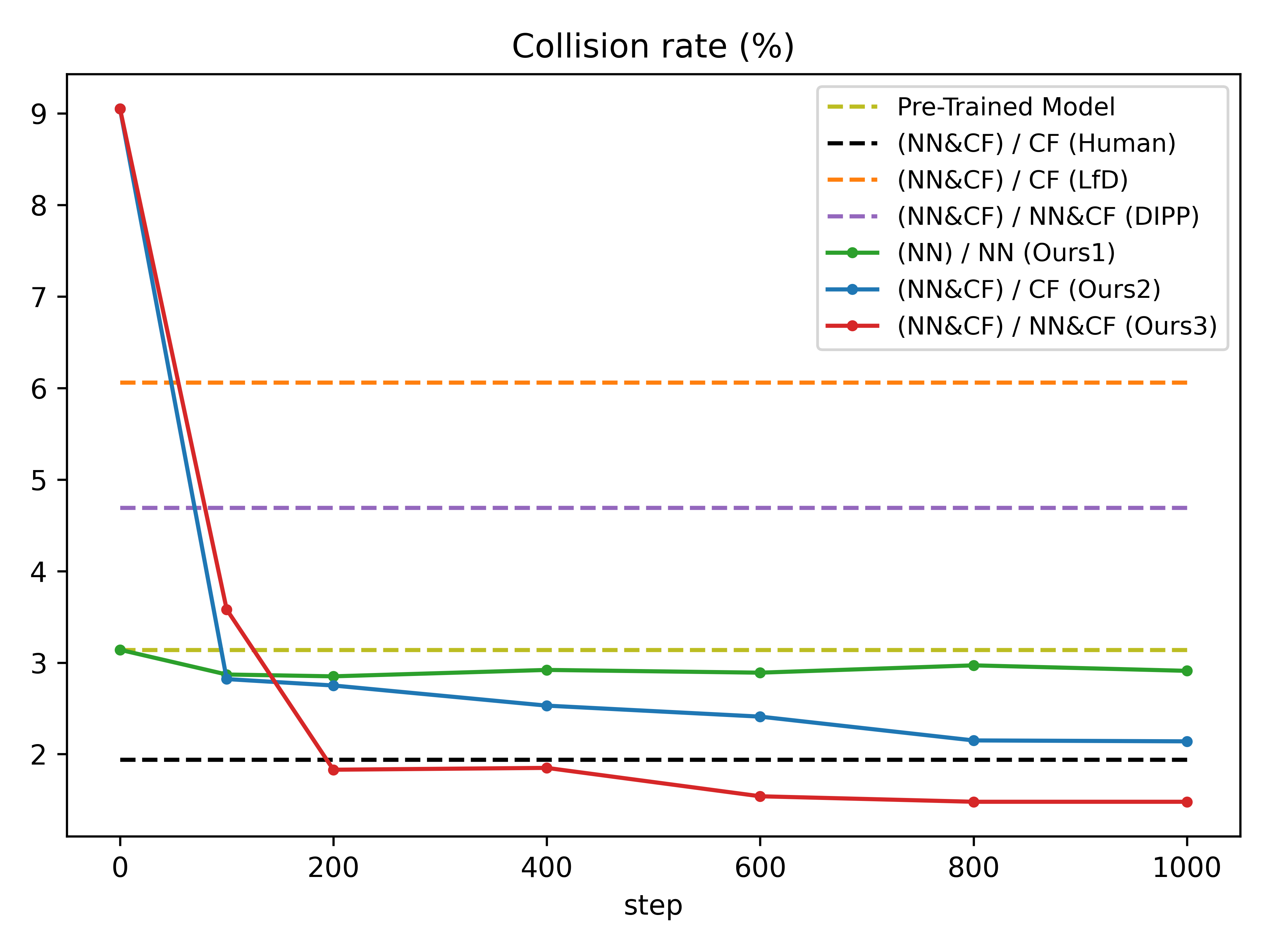}     
    % } % \label{feature_flow_global}  0.335
    \vspace{-0.5cm}
    \caption{Comparison of different steps. The figure depicts the average values from three repeated experiments. We perform 1000 times parameter updates for each model, reducing the learning rate by half every 200 steps. The parentheses indicate the fine-tuning framework utilized, with the outer parentheses denoting the part for parameter update.
    }
    % \vspace{-0.8cm}
    \label{fig:step_results}
\end{figure}

\textbf{Comparison of different steps.} To further evaluate whether our algorithm is prone to overfitting, we present in Fig. \ref{fig:step_results} the trends of style feature matching error and collision rate changes with the update steps across three learning structures. The curve values in the figure represent the average results obtained from three out-of-domain experiments. 
Due to overfitting to sparse user data, ``LfD" and ``DIPP" exhibit higher collision rates and poorer style generalization capability. 
Upon analyzing the curve trends, the final convergence outcome of the ``NN\&CF" framework surpasses that of the ``NN" framework in two key metrics: style error and collision rate. ``(NN\&CF) / CF" initially converges faster. However, since the cost function solely learns linear weights, its final convergence performance is limited. ``(NN\&CF) / NN\&CF" proves less effective when the number of update steps is small. In such cases, the linear weight of the cost function hasn't been adequately learned, and it hasn't adapted well to the neural network. However, with sufficient update steps, ``(NN\&CF) / NN\&CF" achieves the best performance and even outperforms manual adjustments. This is attributed to the strong fitting ability of the neural network, which mitigates the problem of the cost function's insufficient fitting capacity.

\subsection{Ablation Study}
% 消融实验
\begin{table}[!htp] 
% \vspace{-0.1cm}
% \hspace{-0.2cm}
% \renewcommand{\arraystretch}{1.25}
\caption{Loss ablation study}
\vspace{-0.2cm}
\hspace{-0.2cm}
\scalebox{0.85}{
\renewcommand{\arraystretch}{1.30}
\setlength{\tabcolsep}{2.5pt}{
\centering
\begin{threeparttable}
\begin{tabular}{@{}c|>{\columncolor{lightgray}} c|c|cccc@{}}
\toprule
% \multirow{2}{*}{\makecell[c]{Fine-Tuning\\(Structure) Method}} 
% &\multirow{2}{*}{Loss}&
&&
\multicolumn{1}{c|}{In Domain}&\multicolumn{4}{c}{Out of Domain}\\
\makecell[c]{(Fine-Tuning Structure)\\/ Updated Part} &
Loss &
\multicolumn{1}{c|}{\makecell[c]{Plan error\\@3s ($m$)}} &
% \multicolumn{1}{c|}{\makecell[c]{Style\\error}}&
% \multicolumn{1}{c}{\makecell[c]{Red light\\(\%) ($m$)}} &
% \multicolumn{1}{c}{\makecell[c]{Off route\\(\%) ($m$)}} &
\multicolumn{2}{c|}{\makecell[c]{Prediction error ($m$)\\ ADE@5s FDE@5s}} &
\multicolumn{1}{c}{\makecell[c]{Collision\\(\%)}} &
\multicolumn{1}{c}{\makecell[c]{Style\\error}} \\
\midrule
\multirow{3}{*}{(NN) / NN}
&$\mathcal{L}_{IL}$& 0.4814&  0.6735 & 1.6510& 3.37&0.06291\\
&$\mathcal{L}_{TIL}^{1}$& \textbf{0.4740}&  0.6739& 1.6524& 3.23&0.05551\\
&$\mathcal{L}_{TIL}^{10}$& 0.4792&  \textbf{0.6735}& \textbf{1.6509}& 3.03&0.04640\\
&$\mathcal{L}_{TIL}^{100}$& 0.5076&  0.6753& 1.6579& 2.87&0.04448\\
&$\mathcal{L}_{TIL}^{1000}$& 0.5727&  0.6799& 1.6793& \textbf{2.67}&0.04144\\
&$\mathcal{L}_{IRL}$& 0.8023&  0.7050 & 1.7285& 5.20&\textbf{0.03869}\\
\midrule
% \multirow{3}{*}{(NN\&CF) / NN} % \makecell[c]{(NN\&CF) / NN \\ + (CF / Human)}
% &$\mathcal{L}_{IL}$& 0.8240&  0.6860& 1.6691& 1.79&0.03339\\
% &$\mathcal{L}_{IRL}$& 0.9378&  0.8001& 1.8515& 2.22&0.03324\\
% &$\mathcal{L}_{FSIL}^{1}$& --&  --& --& --&--\\
% &$\mathcal{L}_{FSIL}^{10}$& --&  --& --& --&--\\
% &$\mathcal{L}_{FSIL}^{100}$& 0.8455&  0.6865& 1.6682& 1.77&0.03342\\
% &$\mathcal{L}_{FSIL}^{1000}$& --&  --& --& --&--\\
% \midrule
% \multirow{3}{*}{(NN\&CF) / CF} % \makecell[c]{(NN\&CF) / NN \\ + (CF / Human)}
% &$\mathcal{L}_{IL}$& 0.7141&  --& --& 4.07&0.04727\\
% &$\mathcal{L}_{IRL}$& 1.4079&  --& --& 6.12&0.04083\\
% &$\mathcal{L}_{FSIL}^{1}$& 0.7171&  --& --& 5.12&0.04038\\
% &$\mathcal{L}_{FSIL}^{10}$& 0.7700&  --& --& 4.40&0.04087\\
% &$\mathcal{L}_{FSIL}^{100}$& 0.9494&  --& --& 2.82&0.04247\\
% &$\mathcal{L}_{FSIL}^{1000}$& --&  --& --& --&--\\
% \midrule
\multirow{3}{*}{(NN\&CF) / NN\&CF}
&$\mathcal{L}_{IL}$& \textbf{0.6903}&  0.6836& \textbf{1.6669}& 3.88&0.04003\\
&$\mathcal{L}_{TIL}^{1}$& 0.8006&  0.6949& 1.6816& 5.57&0.03766\\
&$\mathcal{L}_{TIL}^{10}$& 0.7604&  \textbf{0.6837}& 1.6714& 5.50&0.03800\\
&$\mathcal{L}_{TIL}^{100}$& 0.7813&  \textbf{0.6837}& 1.6756& \textbf{3.58}&\textbf{0.03585}\\
&$\mathcal{L}_{TIL}^{1000}$& 1.1904&  0.6862& 1.6788& 4.85&0.04366\\
&$\mathcal{L}_{IRL}$& 1.4948&  0.7109& 1.7146& 5.46&0.05056\\
\bottomrule
\end{tabular}
\begin{tablenotes}
\footnotesize
\item NN: neural network, CF: cost function, NN\&CF: joint neural network and cost function.
\end{tablenotes}
\end{threeparttable}
\label{table:ablation_loss}
    } 
}
\vspace{-0.2cm}
\end{table}
\textbf{Loss ablation study.} As depicted in Table \ref{table:ablation_loss}, various losses are ablated. The input batch data for all experiments in the table consists of a 50\% mixture of expert data and user data. We conduct 100 updates for each setting and present the averages of three repeated experiments. To better validate the efficacy of the loss, we only modify the loss to explore the effectiveness of the proposed method. The results indicate that the method utilizing only $\mathcal{L}_{IL}$ for parameter updates exhibits the poorest performance in the ``style error" indicator and achieves the lowest approximation accuracy for user style. Conversely, the method employing only $\mathcal{L}_{IRL}$ performs inadequately on other vital planning performance metrics. $\mathcal{L}_{TIL}$ effectively balances user style approximation with the performance enhancements of the planning and forecasting modules. Additionally, we conduct parametric sensitivity experiments on the weight proportion between the $\mathcal{L}_{IL}$ and $\mathcal{L}_{IRL}$. The results indicate that $\alpha=100$ serves as a suitable trade-off between planned performance improvements and user style approximation, which tends to result in lower style error and collision rate.
\begin{table}[!htp]
\vspace{-0.2cm}
\hspace{-0.3cm}
\renewcommand{\arraystretch}{1.20}
\caption{Mixed proportion ablation study}
\hspace{-0.1cm}
% \caption{The inference speed and the prediction performance of models on the Argoverse validation set.} 
\scalebox{0.80}{
\setlength{\tabcolsep}{2.5pt}{
\centering
\begin{threeparttable}
\begin{tabular}{@{}c|>{\columncolor{lightgray}}c|c|cccc@{}}
\toprule
&&
\multicolumn{1}{c|}{In Domain}&\multicolumn{4}{c}{Out of Domain}\\
\makecell[c]{(Fine-Tuning Structure)\\/ Updated Part}&
% \makecell[c]{The Proportion\\of Expert Data}&
\makecell[c]{Proportion}&
\multicolumn{1}{c|}{\makecell[c]{Plan error\\@3s ($m$)}} &
% \multicolumn{1}{c|}{\makecell[c]{IRL\\error}}&
% \multicolumn{1}{c}{\makecell[c]{Red light\\(\%) ($m$)}} &
% \multicolumn{1}{c}{\makecell[c]{Off route\\(\%) ($m$)}} &
\multicolumn{2}{c|}{\makecell[c]{Prediction error ($m$)\\ ADE@5s FDE@5s}} &
\multicolumn{1}{c}{\makecell[c]{Collision\\(\%)}} &
\multicolumn{1}{c}{\makecell[c]{Style\\error}} \\
\midrule
\multirow{5}{*}{(NN) / NN}
&$0\%$& 0.5237&  0.7081& 1.7226& 3.61&0.05684\\
&$25\%$& 0.5086&  0.6835& 1.6723& 2.98&0.04993\\
&$50\%$& \textbf{0.5076}&  0.6753& 1.6579& \textbf{2.87}&0.04448\\
&$75\%$& 0.5480&  \textbf{0.6654}& \textbf{1.6452}& 2.95&\textbf{0.04301}\\
&$100\%$& 0.5786&  0.6677& 1.6473& 2.98&0.04488\\
\midrule
% \multirow{5}{*}{(NN\&CF) / NN}
% &$0\%$& 0.5804&  0.7170& 1.7264& 3.75&0.06357\\
% &$25\%$& 0.8401&  0.6922& 1.6837& 1.78&0.03327\\
% &$50\%$& 0.8455&  0.6865& 1.6682& 1.77&0.03342\\
% &$75\%$& 0.8405&  0.6824& 1.6613& 1.77&0.03347\\
% &$100\%$& 0.8138&  0.6750& 1.6598& 1.74&0.03307\\
% \midrule
% \multirow{5}{*}{(NN\&CF) / CF}
% &$0\%$& 1.1915&  --& --& 6.74&0.03923\\
% &$25\%$& --&  --& --& --&--\\
% &$50\%$& 0.9494&  --& --& 2.82&0.04247\\
% &$75\%$& --&  --& --& --&--\\
% &$100\%$& --&  --& --& --&--\\
% \midrule
\multirow{5}{*}{(NN\&CF) / NN\&CF}
&$0\%$& \textbf{0.5743}&  0.7175& 1.7359& 3.82&0.06404\\
&$25\%$& 0.7509&  0.6889& 1.6793& 3.08&0.03769\\
&$50\%$& 0.7813&  0.6837& 1.6756& 3.58&\textbf{0.03585}\\
&$75\%$& 0.7287&  \textbf{0.6762}& \textbf{1.6581}& \textbf{2.93}&0.03636\\
&$100\%$& 0.8459&  0.6885& 1.6764& 3.61&0.04460\\
\bottomrule
\end{tabular}
\begin{tablenotes}
\footnotesize
\item NN: neural network, CF: cost function, NN\&CF: joint neural network and cost function, Proportion: the proportion of expert data in the mini-batch of input.
\end{tablenotes}
\end{threeparttable}
\label{table:ablation_proportion}
    } 
}
\vspace{-0.5cm}
\end{table}

\textbf{Mixed proportion ablation study.} As presented in Table \ref{table:ablation_proportion}, we examine different ratios of mixing expert data to explore their influence on fine-tuning outcomes. For each setting, we conduct 100 updates and report the average of three repeated experiments. Throughout the experiment, we utilize $\mathcal{L}_{TIL}$ to update the parameters and set the $\alpha$ parameter to 100. The results indicate that when the mixing ratio of expert data is 0\%, the model tends to overfit to user data, resulting in poor generalization performance since only user data is utilized. Conversely, when the mixing ratio of expert data is 100\%, the planning error within the user domain increases significantly because only expert data is input. Across experiments, it is evident that the trade-off between planning performance enhancement and user style approximation is better achieved with a mixture ratio of 75\%, often resulting in lower style error and collision rate.

\subsection{Qualitative Results}
\textbf{Visualization of the final results.} We utilized the three models obtained from the Fig. \ref{fig:step_results} experiment to conduct further analysis, presenting the qualitative findings in Fig. \ref{fig:visual_compare}. The figure showcases visualizations of the three models. The trajectory duration is fixed at 5 seconds. The purple curve denotes the predicted trajectory of other cars, with the model considering only the ten nearest other cars for trajectory prediction. 

\section{Conclusion}
In this paper, we propose an instance-based transfer imitation learning approach aimed at addressing the challenge of scarcity in user domain data. We employ a pre-trained fine-tuning framework to transfer expertise from the expert domain to alleviate the data scarcity issue in the user domain. Our experimental results demonstrate that our method achieves competitive planning performance while effectively capturing the driving style of the user. 
At the same time, it is vital to acknowledge the limitations of this work. Firstly, we do not conduct closed-loop real-world experiments. Besides, we utilize the error of the feature expectation of the trajectory to measure user style. To better represent user driving styles or cater to user needs, it is essential to develop more appropriate methods for measuring user styles.

\bibliographystyle{IEEEtran}
\bibliography{main}

\end{document}